\documentclass[11pt]{article}
\usepackage[margin=1in]{geometry}
\usepackage{amsmath,amssymb}
\usepackage{graphicx}
\usepackage{booktabs}
\usepackage{enumitem}
\usepackage{hyperref}
\usepackage{url}
\usepackage{natbib}

\providecommand{\checkmark}{\ensuremath{\surd}}
\providecommand{\texttimes}{\ensuremath{\times}}

\title{Information Routing in Atomistic Foundation Models:\\
How Task Alignment and Equivariance Shape Linear Disentanglement}

\author{
  Joshua Steier\footnote{Correspondence: \texttt{joshuasteier@gmail.com}} \\
  Independent Researcher
}

\begin{document}
\maketitle


\begin{abstract}
What determines whether a molecular property prediction model organizes its
representations so that geometric and compositional information can be
cleanly separated? We introduce Compositional Probe Decomposition (CPD),
which linearly projects out composition signal and measures how much
geometric information remains accessible to a Ridge probe. We validate CPD
with four independent checks, including a structural isomer benchmark where
compositional projections score at chance while geometric residuals reach
94.6\% pairwise classification accuracy.

Across ten models from five architectural families on QM9, we find a
\emph{linear accessibility gradient}: models differ by $6.6\times$ in
geometric information accessible after composition removal
($R^2_{\mathrm{geom}}$ from 0.081 to 0.533 for HOMO-LUMO gap). Three
factors explain this gradient. Task alignment dominates: models trained on
HOMO-LUMO gap ($R^2_{\mathrm{geom}}$ 0.44--0.53) outscore energy-trained
models by $\sim$0.25 $R^2$ regardless of architecture. Within-architecture
ablations on two independent architectures confirm this: PaiNN drops from
0.53 to 0.31 when retrained on energy, and MACE drops from 0.44 to 0.08.
Data diversity
partially compensates for misaligned objectives, with MACE pretrained on
MPTraj (0.36) outperforming QM9-only energy models.

Inside MACE's representations, information routes by symmetry type: $L{=}1$
(vector) channels preferentially encode dipole moment ($R^2 = 0.59$ vs.\
0.38 in $L{=}0$), while $L{=}0$ (scalar) channels encode HOMO-LUMO gap
($R^2 = 0.76$ vs.\ 0.34 in $L{=}1$). This pattern is absent in ViSNet. We
also show that nonlinear probes produce misleading results on residualized
representations, recovering $R^2 = 0.68$--$0.95$ on a purely compositional
target, and recommend linear probes for this setting.
\end{abstract}


\section{Introduction}

Atomistic foundation models such as MACE \citep{batatia2022mace}, SchNet
\citep{schutt2017schnet}, DimeNet++ \citep{klicpera2020dimenet}, PaiNN
\citep{schutt2021painn}, and ViSNet \citep{wang2024visnet} now predict
molecular energies, forces, and electronic properties with accuracy
approaching density functional theory at a fraction of the cost. As these
models are deployed for virtual screening, catalyst design, and molecular
dynamics, a basic question remains open: what do their intermediate
representations encode, and how is that information organized? If a
representation cleanly separates what a molecule is made of (its
composition) from how those atoms are arranged (its geometry), downstream
tasks can selectively access the relevant signal. A model that entangles
these factors forces every application to disentangle them from scratch.

Representational probing offers a natural way to answer this question:
train a simple predictor on internal activations and measure what it
recovers \citep{conneau2018probing, hewitt2019structural,
alain2017understanding}. But molecular properties reflect both composition
and geometry, and these contributions are correlated. A probe on raw
representations cannot distinguish genuine geometric encoding from
compositional shortcuts. To isolate geometric signal, the composition
contribution must first be removed.

We initially attempted this removal with Ridge regression and probed the
residuals using gradient boosted trees (GBTs). The results were striking
and wrong: on a purely compositional target (average atomic mass), GBT
probes recovered $R^2 = 0.68$--$0.95$ from residuals that should contain
no relevant signal. GBTs exploit high-dimensional residuals to reconstruct
the projected-out signal, producing systematically inflated scores. This
failure motivated both the methodological core of this paper and the
exclusive use of linear probes for residualized representations.

We introduce Compositional Probe Decomposition (CPD), which fits an OLS
projection to remove composition signal within each cross-validation fold,
then probes the residual with Ridge regression to quantify linearly
accessible geometric information. We validate CPD with four independent
checks, including a structural isomer benchmark where compositional
projections score at chance while geometric residuals reach 94.6\% pairwise
classification accuracy.

Applying CPD to ten models from five architectural families on QM9
(small organic molecules with up to 9 heavy atoms), we find a
\emph{linear accessibility gradient}: a $6.6\times$ spread in geometric
information accessible after composition removal. This gradient is shaped
by three interacting factors whose relative importance overturns a natural
assumption in the field. Section~\ref{sec:gradient} presents the gradient
and decomposes it; Section~\ref{sec:routing} reveals structured information
routing through MACE's equivariant channels; and
Section~\ref{sec:inflation} demonstrates that nonlinear probes produce
systematically inflated scores on residualized representations.

\paragraph{Contributions.}
\begin{enumerate}
    \item \textbf{CPD as a validated probing methodology}, including a
    structural isomer ground-truth test and evidence that nonlinear probes
    produce systematically inflated scores on residualized representations.

    \item \textbf{A three-factor linear accessibility gradient} across
    ten models, showing task alignment ($\sim$0.25 $R^2$ gap) dominates
    over equivariance and data diversity. Within-architecture ablations on
    two independent architectures (PaiNN: $\Delta = 0.223$; MACE:
    $\Delta = 0.338$) confirm that the training objective, not
    architecture, drives linear geometric accessibility.

    \item \textbf{Information routing by irreducible representation} in
    MACE, where scalar and vector channels preferentially encode properties
    matching their symmetry type.

    \item \textbf{Robustness across twelve checks}, with model rankings
    perfectly preserved (Spearman $\rho = 1.0$) under concept erasure,
    alternative composition features, partial regression, and isomer
    classification.
\end{enumerate}


\section{Background and Related Work}

This work sits at the intersection of three lines of research:
representational probing, equivariant neural networks for molecular
modeling, and disentangled representation learning. We briefly review each
and identify the gap that CPD addresses.

\subsection{Representational Probing}

The idea of training a simple classifier or regressor on a neural network's
internal activations to understand what it has learned originates in natural
language processing. \citet{alain2017understanding} trained linear
classifiers on intermediate layers of deep networks to track how
representations evolve with depth. \citet{conneau2018probing} introduced a
suite of ``probing tasks'' for sentence embeddings, measuring whether
syntactic and semantic features could be recovered from fixed
representations. \citet{hewitt2019structural} showed that syntactic tree
distances are approximately encoded as linear maps in BERT's embedding
space, establishing that some language models organize structural
information in a linearly accessible way.

A key methodological lesson from this literature is that probe complexity
matters. \citet{hewitt2019control} demonstrated that sufficiently expressive
probes can achieve high accuracy even on random representations, and
proposed control tasks to calibrate probe selectivity. We encounter an
analogous problem in the molecular setting: gradient boosted tree probes
recover high $R^2$ from residualized representations even when the target
is purely compositional, precisely because their expressiveness allows them
to reconstruct removed signal.

Probing has seen limited application in atomistic ML.
\citet{batatia2022mace_analysis} analyzed MACE representations but did not
separate composition from geometry. To our knowledge, no prior work has
systematically probed molecular representations after removing the
composition confound, which is the central methodological contribution of
this paper.

\subsection{Equivariant Architectures for Molecular Property Prediction}

Neural networks for molecular property prediction have evolved along two
axes: how they represent atomic environments and whether they respect
physical symmetries. SchNet \citep{schutt2017schnet} introduced continuous
filter convolutions on interatomic distances, producing representations
that are invariant to rotation and translation by construction. DimeNet++
\citep{klicpera2020dimenet} incorporated bond angles through directional
message passing, adding sensitivity to local geometry while maintaining
global invariance.

A second generation of models explicitly encodes equivariance, meaning
their intermediate representations transform predictably under rotations.
PaiNN \citep{schutt2021painn} maintains separate scalar and vector
channels, where vector features rotate with the coordinate frame. MACE
\citep{batatia2022mace} goes further, constructing messages using tensor
products of spherical harmonics, which produces features that transform as
irreducible representations of SO(3) at multiple angular momentum orders
($L{=}0, 1, 2, \ldots$). ViSNet \citep{wang2024visnet} takes a different
approach, using runtime geometric computations on scalar and vector
features rather than explicit tensor product operations. ANI-2x
\citep{devereux2020ani2x} uses handcrafted symmetry functions that are
invariant by design, with no learned equivariant features.

A natural hypothesis is that models with richer equivariant structure
produce representations where geometric information is more accessible.
Our results partially support and partially refute this hypothesis: the
relationship between equivariance and geometric accessibility depends
critically on what the model was trained to predict.

\subsection{Disentangled Representations}

Disentangled representation learning seeks to produce latent spaces where
distinct generative factors are encoded in separate dimensions or subspaces
\citep{bengio2013representation, higgins2017betavae}. \citet{locatello2019}
showed that unsupervised disentanglement is impossible without inductive
biases, motivating the study of which architectural and training choices
encourage separation of factors. More recently, \citet{huh2024platonic}
proposed the Platonic Representation Hypothesis: that models trained on
different data and with different architectures converge toward a shared
statistical model of reality. Our results complicate this picture. While all
ten models encode both composition and geometry, they differ dramatically
in \emph{how} that information is organized, with task alignment and
architecture interacting to produce qualitatively different representation
structures.

In the molecular domain, the relevant factors are composition (which
elements are present and in what proportions) and geometry (how those atoms
are arranged in space). These factors are not independent: composition
constrains the space of possible geometries, and many molecular properties
correlate with both. This correlation is what makes naive probing
misleading and what CPD is designed to address.

Our notion of disentanglement is specific and operational. We say a
representation is \emph{linearly disentangled} with respect to composition
and geometry if, after linearly projecting out composition signal, a linear
probe can still recover geometric information. This differs from the
$\beta$-VAE tradition \citep{higgins2017betavae} in that we measure
accessibility of a known factor (geometry) after removal of another known
factor (composition), rather than seeking to discover latent factors. It is
closer in spirit to the concept erasure framework of
\citet{belrose2023leace}, which finds the optimal linear guard against
concept leakage. We use LEACE as one of our four validation checks and show
that it produces rankings identical to CPD (Spearman $\rho = 1.0$ for three
of four target properties).

\subsection{Composition as a Confound in Molecular ML}

The observation that composition alone is a strong predictor of molecular
properties is well established. Matminer \citep{ward2018matminer} and
CrabNet \citep{wang2021crabnet} achieve competitive performance on
materials property benchmarks using only compositional descriptors. On QM9,
simple composition features (element fractions and atom count) explain
roughly 39\% of HOMO-LUMO gap variance and 87\% of polarizability variance
under linear regression.

This means that any model achieving high $R^2$ on these targets may be
relying partly or entirely on composition. Probing studies that report high
$R^2$ without controlling for composition cannot distinguish geometric
encoding from compositional memorization. CPD addresses this directly by
removing the linear composition subspace before probing, and the structural
isomer benchmark provides a composition-free evaluation where only
geometric information can contribute to prediction.


\section{Methodology}

We describe CPD in three parts: the composition features that define the
signal to be removed (\S\ref{sec:comp_features}), the projection that
removes it (\S\ref{sec:projection}), and the probing protocol that
measures what remains (\S\ref{sec:probing}).

\subsection{Composition Features}
\label{sec:comp_features}

For each molecule, we construct a composition vector $\mathbf{z} \in
\mathbb{R}^k$ encoding what elements are present and in what proportions.
Our default specification uses $k = 6$ features: element fractions for C,
H, N, O, and F (each divided by the total atom count), plus a standardized
atom count (z-scored across the dataset). Because the element fractions sum
to approximately one, all downstream regressions include an intercept term
to avoid rank deficiency.

A natural concern is that results might depend on how composition is
defined. We test this by repeating all analyses under three alternative
specifications: raw element counts with raw atom count (unnormalized),
element fractions without atom count, and binary element presence indicators
with standardized atom count. Section~\ref{sec:robustness} shows the model
ranking is invariant to this choice (Spearman $\rho = 1.0$ across all
pairwise comparisons).

\subsection{Compositional Probe Decomposition}
\label{sec:projection}

Let $\mathbf{X} \in \mathbb{R}^{n \times d}$ be the matrix of molecular
representations extracted from a frozen model, and let $\mathbf{Z} \in
\mathbb{R}^{n \times k}$ be the corresponding composition feature matrix.
CPD removes the linear relationship between composition and representation
by fitting an ordinary least squares regression of $\mathbf{X}$ on
$\mathbf{Z}$ and taking the residual:
\begin{equation}
    \hat{\boldsymbol{\beta}} = (\mathbf{Z}^\top \mathbf{Z})^{-1}
    \mathbf{Z}^\top \mathbf{X}, \qquad
    \mathbf{X}_{\mathrm{geom}} = \mathbf{X} -
    \mathbf{Z}\hat{\boldsymbol{\beta}}.
    \label{eq:cpd}
\end{equation}
The residual $\mathbf{X}_{\mathrm{geom}}$ is the component of the
representation that is linearly orthogonal to composition. The
compositional component is $\mathbf{X}_{\mathrm{comp}} =
\mathbf{Z}\hat{\boldsymbol{\beta}}$.

\begin{figure}[t]
\centering
\includegraphics[width=\textwidth]{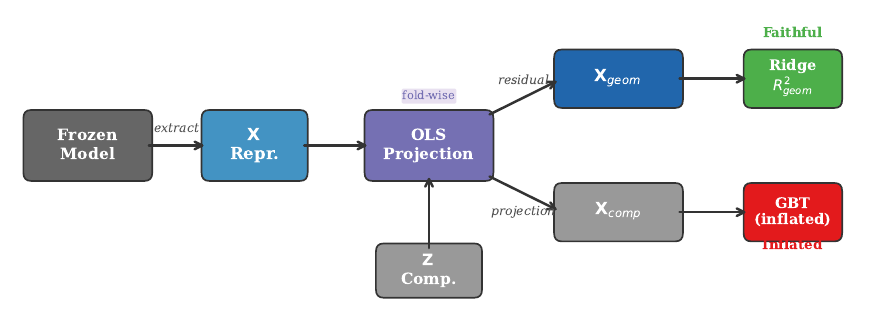}
\caption{The CPD pipeline. Representations $\mathbf{X}$ are extracted from
a frozen model, then OLS projection (fold-wise) removes the linear
composition component $\mathbf{Z}\hat{\boldsymbol{\beta}}$, producing a
geometric residual $\mathbf{X}_{\mathrm{geom}}$. Ridge regression on the
residual yields $R^2_{\mathrm{geom}}$ (faithful). Gradient boosted trees
on the same residual produce inflated scores by reconstructing the
projected-out composition signal.}
\label{fig:pipeline}
\end{figure}

This operation is equivalent to QR projection: computing $\mathbf{Q},
\mathbf{R} = \mathrm{qr}(\mathbf{Z})$ and setting
$\mathbf{X}_{\mathrm{geom}} = \mathbf{X} - \mathbf{Q}\mathbf{Q}^\top
\mathbf{X}$. We use the OLS formulation because it extends naturally to the
fold-wise setting described below. Figure~\ref{fig:pipeline} illustrates
the full CPD pipeline.

\paragraph{Fold-wise projection.} Computing
$\hat{\boldsymbol{\beta}}$ on the full dataset and then evaluating probes
under cross-validation introduces a subtle information leak: the projection
coefficients are informed by test-fold composition vectors. While the leak
is small for low-dimensional $\mathbf{Z}$ ($k = 6$), we eliminate it
entirely by fitting $\hat{\boldsymbol{\beta}}$ on training data only within
each fold and applying the resulting projection to both train and test
splits. This is equivalent to the Frisch-Waugh-Lovell theorem applied
fold-wise, and it ensures that the test-fold residuals are computed from a
projection the model has not seen. In practice, fold-wise CPD produces
slightly higher $R^2_{\mathrm{geom}}$ values than global CPD (mean
$\Delta = +0.04$), because global projection removes a wider subspace. The
model ranking is identical under both approaches.

\paragraph{What the residual contains.} We use the notation
$R^2_{\mathrm{geom}}$ for readability, but the residual
$\mathbf{X}_{\mathrm{geom}}$ is not purely three-dimensional geometry. It
captures all information that is linearly independent of composition,
including molecular topology, bond connectivity, conformational features,
and any nonlinear interactions between composition and structure. We use
``non-compositional signal'' and ``geometric information'' interchangeably
throughout, with the understanding that the latter is a convenient
shorthand for the former.

\subsection{Probing Protocol}
\label{sec:probing}

We extract molecular representations from each frozen, pretrained model by
passing a fixed set of 2{,}000 QM9 molecules through the network and saving
the final-layer activations as mean-pooled per-molecule vectors. The 2{,}000
molecules are selected by a fixed random permutation (seed 42) to ensure
identical evaluation sets across all models.

For each model, we train Ridge regression probes (\texttt{RidgeCV} with 20
log-spaced regularization values from $10^{-3}$ to $10^{6}$,
\texttt{fit\_intercept=True}) on four targets: HOMO-LUMO gap, dipole
moment, polarizability, and zero-point vibrational energy (ZPVE). All
results use 5-fold cross-validation repeated across 30 random seeds, with
reported values being the mean $R^2 \pm$ standard deviation across seeds.

\paragraph{Why linear probes.} A central finding of this paper is that
nonlinear probes produce misleading results on residualized
representations. Gradient boosted trees recover $R^2 = 0.68$--$0.95$ on
average atomic mass (a purely compositional target) from
$\mathbf{X}_{\mathrm{geom}}$, which should contain no compositional signal.
This inflation occurs because GBTs have enough capacity to reconstruct the
projected-out composition signal from high-dimensional residuals through
nonlinear feature interactions. Linear probes cannot perform this
reconstruction and therefore provide a faithful measure of what is linearly
accessible in the residual. We report GBT results only as a cautionary
comparison.

\subsection{Models}
\label{sec:models}

We evaluate ten models spanning five architectural families
(Table~\ref{tab:models}). The models differ in three dimensions that our
analysis aims to disentangle: equivariant structure, training objective, and
training data.

\begin{table}[t]
\centering
\caption{Models evaluated. ``Equiv.'' indicates whether the architecture
produces equivariant intermediate representations. ``TP'' indicates tensor
product message passing. PaiNN and MACE each appear in both HOMO-LUMO and
energy-trained variants, enabling two independent $2 \times 2$ factorial
tests of task alignment.}
\label{tab:models}
\small
\begin{tabular}{lcccll}
\toprule
\textbf{Model} & \textbf{Equiv.} & \textbf{TP} & \textbf{Params} &
\textbf{Training Objective} & \textbf{Training Data} \\
\midrule
PaiNN           & \checkmark & \texttimes & 1.17M & HL gap        & QM9 (30 ep) \\
ViSNet          & \checkmark & \texttimes & 1.67M & HL gap        & QM9 (50 ep) \\
MACE-HL         & \checkmark & \checkmark & 0.73M & HL gap        & QM9 (30 ep) \\
MACE pretrained & \checkmark & \checkmark & 3.70M$^*$ & E + F     & MPTraj (millions) \\
ANI-2x          & \texttimes & \texttimes & 13.7M & E             & ANI-1x (diverse) \\
PaiNN-energy    & \checkmark & \texttimes & 1.17M & E (U$_0$)     & QM9 (30 ep) \\
SchNet          & \texttimes & \texttimes & 0.46M & E (U$_0$)     & QM9 (30 ep) \\
DimeNet++       & \texttimes & \texttimes & 1.89M & E (U$_0$)     & QM9 (30 ep) \\
MACE QM9 97ep   & \checkmark & \checkmark & 0.73M & E             & QM9 (97 ep) \\
MACE QM9 30ep   & \checkmark & \checkmark & 0.73M & E             & QM9 (30 ep) \\
\bottomrule
\multicolumn{6}{l}{\footnotesize $^*$Published value for MACE-MP-0-medium.}
\end{tabular}
\end{table}

PaiNN, ViSNet, and MACE-HL are trained on HOMO-LUMO gap, the same property
we use as our primary probe target. This is deliberate: it allows us to
test whether task alignment between training and probing objectives affects
disentanglement. PaiNN-energy uses identical architecture to PaiNN but is
trained on QM9 energy (U$_0$), and MACE-HL uses identical architecture to
the MACE QM9 variants but is trained on HOMO-LUMO gap. Together these
provide two independent within-architecture ablations of training objective.
The remaining models are all trained on energy-related objectives, providing
a within-objective comparison group where architecture and data vary but
the training target does not.

The MACE family provides a controlled ablation. MACE pretrained, MACE QM9
97ep, and MACE QM9 30ep share identical architecture (ScaleShiftMACE, 128
scalar + 128 vector channels, $\ell_{\max} = 2$, correlation order 3) but
differ in training data and duration. Any difference in their
$R^2_{\mathrm{geom}}$ scores is attributable to training rather than
architecture. Parameter counts range from 0.46M (SchNet) to 13.7M (ANI-2x)
with no correlation to $R^2_{\mathrm{geom}}$: ANI-2x (13.7M) scores 0.331,
while PaiNN ($12\times$ smaller, 1.17M) scores 0.533. This rules out model
capacity as a confound for the gradient.

\paragraph{Limitations of the model set.} DimeNet++ was trained for only 30
epochs and achieves the lowest $R^2_{\mathrm{full}}$ (0.617), suggesting it
may not have fully converged. Its position on the gradient should be
interpreted with this caveat. We also note that our QM9 model set covers
small organic molecules with up to 9 heavy atoms; the Materials Project
extension (Section~\ref{sec:mp_crystals}) broadens coverage to periodic
crystals. We discuss remaining scope limitations in
Section~\ref{sec:limitations}.

\subsection{Irreducible Representation Analysis}
\label{sec:irreps}

MACE's equivariant architecture produces representations that decompose
into channels transforming as irreducible representations of SO(3). At each
layer, the output contains scalar ($L{=}0$) features that are invariant
under rotation and vector ($L{=}1$) features that rotate with the molecular
frame. This decomposition allows us to ask whether the model routes
different types of information through different symmetry channels.

We extract $L{=}0$ and $L{=}1$ channels separately from MACE's final
interaction layer. The $L{=}1$ channels are three-dimensional vectors; to
probe them for scalar properties, we convert them to invariant features by
computing the norm (magnitude) of each vector. We then run the standard CPD
and Ridge probing pipeline independently on each channel type.

For comparison, we perform the same analysis on ViSNet, which also
maintains scalar and vector feature streams but constructs them through
runtime geometric operations rather than tensor products. If information
routing is a general property of equivariant architectures, both models
should show similar channel specialization. If it is specific to tensor
product message passing, the pattern should appear in MACE but not ViSNet.

\subsection{Structural Isomer Benchmark}
\label{sec:isomers}

The strongest test of whether CPD successfully separates composition from
geometry is to evaluate it on molecules that share identical composition.
Structural isomers have the same molecular formula but different atomic
arrangements, so any difference in their properties is entirely geometric.

Within our 2{,}000-molecule probe set, we identify 150 isomer groups
containing 1{,}931 molecules and forming 38{,}784 isomer pairs. For each
pair $(i, j)$ of isomers, we compute the signed representation difference
$\mathbf{X}[i] - \mathbf{X}[j]$ and train a logistic regression classifier
(5-fold CV) to predict which molecule has the higher HOMO-LUMO gap.

This benchmark has a built-in control: since isomers share identical
composition vectors, the compositional component
$\mathbf{X}_{\mathrm{comp}}$ must produce the same vector for both
molecules, and classification accuracy must be at chance (50\%). Any
accuracy above chance in $\mathbf{X}_{\mathrm{geom}}$ is direct evidence
that the geometric residual carries structural information that composition
cannot explain.


\section{Results}

We organize our results around three findings: the linear accessibility
gradient and the three factors that shape it (\S\ref{sec:gradient}),
information routing through MACE's equivariant channels
(\S\ref{sec:routing}), and systematic inflation from nonlinear probes
(\S\ref{sec:inflation}). Robustness analyses appear in
Section~\ref{sec:robustness}.

\subsection{The Linear Accessibility Gradient}
\label{sec:gradient}

Table~\ref{tab:main} presents the main result. After removing composition
signal via fold-wise CPD, models differ by a factor of $6.6\times$ in how
much geometric information remains linearly accessible. PaiNN achieves the
highest score ($R^2_{\mathrm{geom}} = 0.533$) and MACE QM9 30ep the lowest
(0.081). The compositional component $R^2_{\mathrm{comp}}$ is approximately
constant across models ($\approx 0.39$ for HOMO-LUMO gap), confirming that
it reflects the dataset-level correlation between composition and property
rather than anything model-specific.\footnote{MACE QM9 97ep shows a
slightly lower $R^2_{\mathrm{comp}}$ (0.335) because its representation
space is lower quality overall: the Ridge probe's regularization interacts
differently with a less informative embedding. The composition features
$\mathbf{Z}$ are identical across models; only the representation
$\mathbf{X}$ differs.}

\begin{table}[t]
\centering
\caption{Fold-wise CPD results on HOMO-LUMO gap. All values are mean $R^2
\pm$ standard deviation across 30 seeds $\times$ 5 folds. Models are
ordered by $R^2_{\mathrm{geom}}$.}
\label{tab:main}
\small
\begin{tabular}{lccccc}
\toprule
\textbf{Model} & \textbf{Equiv.} & \textbf{Target} &
$\boldsymbol{R^2_{\mathrm{full}}}$ &
$\boldsymbol{R^2_{\mathrm{geom}}}$ &
$\boldsymbol{R^2_{\mathrm{comp}}}$ \\
\midrule
PaiNN           & \checkmark & HL gap & 0.949 & 0.533 $\pm$ 0.004 & 0.389 \\
ViSNet          & \checkmark & HL gap & 0.877 & 0.513 $\pm$ 0.010 & 0.389 \\
MACE-HL         & \checkmark & HL gap & 0.715 & 0.439 $\pm$ 0.005 & 0.389 \\
MACE pretrained & \checkmark & Energy & 0.786 & 0.364 $\pm$ 0.007 & 0.389 \\
ANI-2x          & \texttimes & Energy & 0.771 & 0.331 $\pm$ 0.006 & 0.389 \\
PaiNN-energy    & \checkmark & Energy & 0.656 & 0.310 $\pm$ 0.005 & 0.389 \\
SchNet          & \texttimes & Energy & 0.692 & 0.262 $\pm$ 0.012 & 0.389 \\
DimeNet++       & \texttimes & Energy & 0.617 & 0.240 $\pm$ 0.005 & 0.389 \\
MACE QM9 97ep   & \checkmark & Energy & 0.289 & 0.101 $\pm$ 0.004 & 0.335$^\dagger$ \\
MACE QM9 30ep   & \checkmark & Energy & 0.342 & 0.081 $\pm$ 0.004 & 0.389 \\
\bottomrule
\multicolumn{6}{l}{\footnotesize $^\dagger$See footnote 1.}
\end{tabular}
\end{table}

\begin{figure}[t]
\centering
\includegraphics[width=\textwidth]{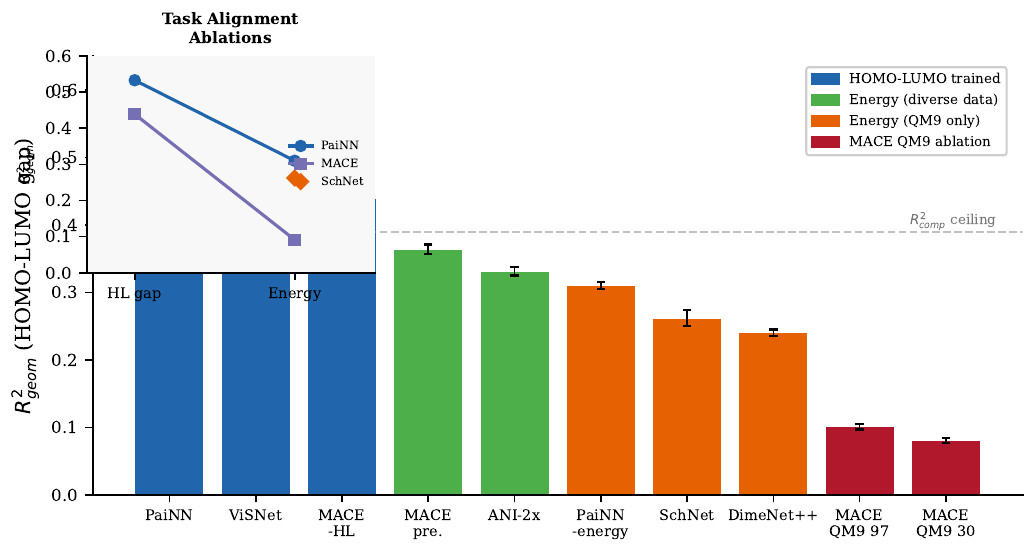}
\caption{The linear accessibility gradient. $R^2_{\mathrm{geom}}$ for
HOMO-LUMO gap across ten models, color-coded by training regime. Three
clusters emerge: HOMO-LUMO-trained (blue), energy-trained with diverse data
(green), and energy-trained on QM9 only (orange/red). Inset: the
$2 \times 2$ factorial crossing PaiNN's architecture with training
objective, showing the task alignment effect ($\Delta = 0.223$) is
$4.6\times$ larger than the architecture effect ($\Delta = 0.048$).
Dashed line shows the composition $R^2$ ceiling.}
\label{fig:gradient}
\end{figure}

The gradient is shaped by three interacting factors, which we now examine.
Figure~\ref{fig:gradient} visualizes the full gradient with models
color-coded by training objective.

\subsubsection{Factor 1: Task Alignment Dominates}

The clearest pattern in Table~\ref{tab:main} is a gap between models
trained on HOMO-LUMO gap and models trained on energy. PaiNN, ViSNet, and
MACE-HL ($R^2_{\mathrm{geom}}$ of 0.533, 0.513, and 0.439 respectively)
outscore every energy-trained model by at least 0.07 $R^2$, and the gap
reaches 0.35 when compared to the energy-trained MACE QM9 variants. This
holds regardless of whether the architecture uses tensor
products.\footnote{MACE-HL's lower $R^2_{\mathrm{full}}$ (0.715 vs.\
PaiNN's 0.949) suggests that the MACE architecture is less sample-efficient
at learning HOMO-LUMO gap from QM9 at 30 epochs, which may suppress its
$R^2_{\mathrm{geom}}$ relative to its ceiling. The rank ordering within the
HL-trained group (PaiNN $>$ ViSNet $>$ MACE-HL) may partly reflect
differences in prediction quality rather than pure representation
organization.}

The interpretation follows directly. HOMO-LUMO gap is sensitive to
molecular geometry (orbital symmetry, conjugation patterns, steric
effects), so models trained on it are forced to develop representations
where geometric signal is prominent. Models trained on total energy face a
different incentive: energy is dominated by composition, with geometry
contributing a smaller correction. Their representations reflect this
balance.

Table~\ref{tab:all_properties} extends this analysis to four properties.
The task-alignment effect is strongest for HOMO-LUMO gap and dipole moment,
where the geometric contribution is large relative to the compositional
floor. For polarizability ($R^2_{\mathrm{comp}} = 0.87$), the composition
ceiling leaves little variance for geometry to explain, and the gradient
compresses accordingly. The PaiNN ablation reveals that task alignment is
property-specific: switching from HOMO-LUMO to energy training drops
$R^2_{\mathrm{geom}}$ for HL gap by 0.223 but \emph{increases} it for ZPVE
by 0.060, because ZPVE correlates more strongly with total energy. The
training objective reshapes which geometric information is linearly
accessible, not just how much.

\begin{table}[t]
\centering
\caption{$R^2_{\mathrm{geom}}$ across four QM9 properties (fold-wise CPD,
30-seed). PaiNN-HL and PaiNN-energy (PaiNN-E) show how task alignment
reshapes geometric accessibility across properties: HOMO-LUMO gap
drops by 0.223 when switching to energy training, while ZPVE increases
by 0.060. Bottom row shows the composition $R^2$ ceiling.}
\label{tab:all_properties}
\small
\begin{tabular}{lcccc}
\toprule
\textbf{Model} & \textbf{HL Gap} & \textbf{Dipole} &
\textbf{Polariz.} & \textbf{ZPVE} \\
\midrule
PaiNN           & 0.533 & 0.171 & 0.051 & 0.188 \\
PaiNN-energy    & 0.310 & 0.201 & 0.081 & 0.248 \\
ViSNet          & 0.513 & 0.141 & 0.056 & 0.190 \\
MACE-HL         & 0.439 & 0.144 & 0.068 & 0.175 \\
MACE pretrained & 0.364 & 0.298 & 0.060 & 0.247 \\
ANI-2x          & 0.331 & 0.266 & 0.063 & 0.247 \\
SchNet          & 0.262 & 0.243 & 0.073 & 0.243 \\
DimeNet++       & 0.240 & 0.201 & 0.074 & 0.224 \\
MACE QM9 97ep   & 0.101 & 0.053 & 0.028 & 0.129 \\
\midrule
$R^2_{\mathrm{comp}}$ & 0.389 & 0.199 & 0.868 & 0.702 \\
\bottomrule
\end{tabular}
\end{table}

\subsubsection{Factor 2: Equivariance Amplifies but Cannot Substitute}

A natural expectation is that equivariant architectures should produce
higher geometric accessibility. Our results show this is conditionally true.

Within the HOMO-LUMO-trained group, PaiNN and ViSNet sit at the top of the
gradient. But within the energy-trained group, equivariance does not help.
MACE QM9 30ep (equivariant) scores 0.081, well below invariant SchNet
(0.262) and DimeNet++ (0.240). This is the most counterintuitive result in
the paper: an equivariant tensor product model, trained on the wrong
objective, produces representations where geometric information is
\emph{less} linearly accessible than in simpler invariant models.

The MACE architecture ablation makes this concrete. MACE pretrained (0.364),
MACE QM9 97ep (0.101), and MACE QM9 30ep (0.081) share identical
architecture. The $4.5\times$ gap between pretrained and 30ep is direct
evidence that training conditions, not tensor product message passing,
determine MACE's position on the gradient. PaiNN reinforces this point from
the opposite direction: it achieves the highest $R^2_{\mathrm{geom}}$ in our
study (0.533) using equivariant message passing \emph{without} tensor
products, demonstrating that tensor products are not necessary for high
geometric accessibility. The critical combination is equivariance plus a
task-aligned training signal.

Two independent within-architecture ablations provide causal tests of task
alignment. PaiNN and PaiNN-energy share identical architecture (6 layers,
128 hidden, 1.17M parameters) but differ only in training objective.
MACE-HL and the MACE QM9 variants share identical architecture
(ScaleShiftMACE, 0.73M parameters) but differ only in training
objective.\footnote{MACE-HL was trained for 30 epochs on QM9 HOMO-LUMO
gap, achieving test RMSE 25.8~meV. See Appendix~\ref{app:mace_training}.}
This yields a $2 \times 2$ factorial replicated across two architectures:

\begin{center}
\small
\begin{tabular}{lccc}
\toprule
& \textbf{HL-trained} & \textbf{Energy-trained} &
\textbf{Task $\Delta$} \\
\midrule
\textbf{PaiNN (equiv, no TP)}  & 0.533 & 0.310 & +0.223 \\
\textbf{MACE (equiv, TP)}      & 0.439 & 0.081--0.101 & +0.338--0.358 \\
\textbf{SchNet (invariant)}    & ---   & 0.262 & --- \\
\bottomrule
\end{tabular}
\end{center}

\noindent Task alignment dominates in both architectures. For PaiNN, the
task effect ($\Delta = 0.223$) is $4.6\times$ larger than the architecture
effect (PaiNN-energy vs.\ SchNet, $\Delta = 0.048$). For MACE, the task
effect is even larger ($\Delta = 0.338$--$0.358$). That task alignment
replicates independently across two architectures with different equivariant
mechanisms (vector channels vs.\ tensor products) is strong evidence that
the training objective, not architecture, is the primary determinant of
linear geometric accessibility.

\subsubsection{Factor 3: Data Diversity Compensates}

Among the energy-trained models, a secondary gradient tracks training data
diversity. MACE pretrained (0.364, MPTraj, millions of structures) and
ANI-2x (0.331, ANI-1x, diverse conformations) both outperform SchNet
(0.262) and DimeNet++ (0.240), which were trained on QM9 alone.

The most informative comparison is MACE pretrained (0.364) versus MACE QM9
97ep (0.101) on identical architecture. The $3.6\times$ gap, with only
training data differing, suggests that exposure to diverse chemical
environments encourages representations that make geometric information
accessible even for properties the model was not trained to predict.

This compensation is partial. Even with massive pretraining diversity, MACE
pretrained (0.364) does not reach the task-aligned cluster (PaiNN/ViSNet at
$\approx$0.52). Data diversity narrows the gap but does not close it.

\subsection{Information Routing by Irreducible Representation}
\label{sec:routing}

MACE's equivariant architecture produces representations with channels
transforming as irreducible representations of SO(3). We test whether
different types of information are routed through channels matching their
physical symmetry by probing $L{=}0$ (scalar) and $L{=}1$ (vector,
norm-converted to invariant features) channels independently.

Table~\ref{tab:irreps} shows the result. Scalar channels carry most of the
HOMO-LUMO gap information ($R^2 = 0.756$, close to the full representation
at 0.786), while vector channels dominate for dipole moment ($R^2 = 0.586$
vs.\ 0.379 in $L{=}0$; Wilcoxon $p = 1.86 \times 10^{-9}$). MACE has
learned to route scalar properties through scalar channels and vector
properties through vector channels.

\begin{table}[t]
\centering
\caption{Information routing in MACE pretrained. Bold indicates the
dominant channel for each property. MACE routes scalar properties through
$L{=}0$ and vector properties through $L{=}1$. Wilcoxon signed-rank test
confirms the dipole preference for $L{=}1$ ($p = 1.86 \times 10^{-9}$,
30 seeds).}
\label{tab:irreps}
\small
\begin{tabular}{lcc}
\toprule
\textbf{Channel} & \textbf{HL Gap $R^2$} & \textbf{Dipole $R^2$} \\
\midrule
Full representation   & 0.786          & 0.499          \\
$L{=}0$ (scalar)      & \textbf{0.756} & 0.379          \\
$L{=}1$ (vector norm) & 0.337          & \textbf{0.586} \\
\bottomrule
\end{tabular}
\end{table}

\begin{figure}[t]
\centering
\includegraphics[width=0.85\textwidth]{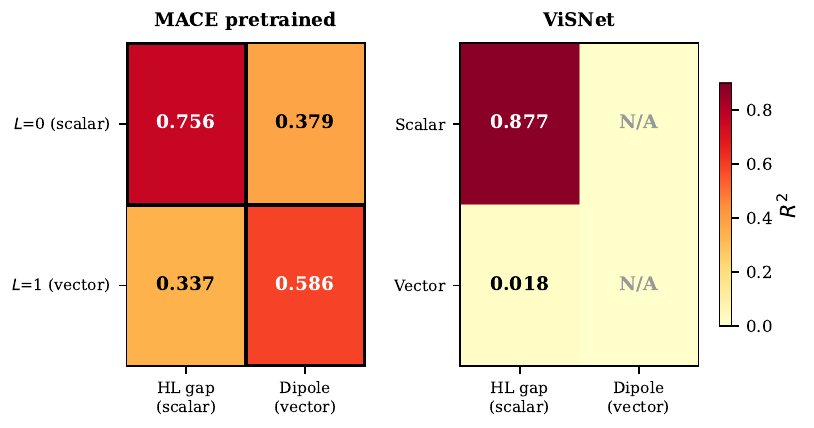}
\caption{Information routing by symmetry channel. Left: MACE routes scalar
properties (HL gap) through $L{=}0$ channels and vector properties (dipole)
through $L{=}1$ channels (black boxes highlight the dominant diagonal).
Right: ViSNet concentrates all information in its scalar stream; vector
channels carry $R^2 = 0.018$.}
\label{fig:irreps}
\end{figure}

For comparison, we probe ViSNet's scalar and vector streams independently
on HOMO-LUMO gap. ViSNet concentrates virtually all linearly accessible
information in its scalar stream ($R^2 = 0.877$), while the vector stream
carries almost nothing ($R^2 = 0.018$). This
suggests that ViSNet's equivariant operations contribute through internal
computation during message passing rather than through persistent structure
in the output. The routing pattern is not a generic consequence of
equivariance but appears specific to architectures that maintain explicit
irreducible representation channels through to the output layer.
Figure~\ref{fig:irreps} visualizes this contrast.

\subsection{Nonlinear Probe Inflation}
\label{sec:inflation}

We motivate the exclusive use of linear probes with a controlled
experiment. We compute $R^2_{\mathrm{geom}}$ using both Ridge regression
and gradient boosted trees on average atomic mass, a purely compositional
target. $\mathbf{X}_{\mathrm{geom}}$ should contain no relevant signal, so
a well-behaved probe should return $R^2 \approx 0$.

Ridge behaves as expected ($R^2_{\mathrm{geom}} \approx -0.003$ across all
models). GBTs recover $R^2 = 0.68$ to $0.95$ from the same residuals
(Table~\ref{tab:inflation}). The explanation is that OLS projection removes
only the \emph{linear} composition component; the residual retains
nonlinear functions of composition that a tree ensemble can exploit to
reconstruct the projected-out target. Linear probes cannot perform this
reconstruction and therefore provide a faithful measure of what is linearly
accessible.

\begin{table}[t]
\centering
\caption{Probe comparison on average atomic mass (purely compositional).
Ridge correctly returns $\approx 0$. GBTs produce inflated scores.}
\label{tab:inflation}
\small
\begin{tabular}{lcc}
\toprule
\textbf{Model} & \textbf{Ridge} & \textbf{GBT} \\
\midrule
MACE pretrained & $-$0.003 & 0.95 \\
ViSNet          & $-$0.003 & 0.89 \\
ANI-2x          & $-$0.002 & 0.93 \\
SchNet          & $-$0.003 & 0.85 \\
DimeNet++       & $-$0.003 & 0.68 \\
MACE QM9        & $-$0.003 & 0.71 \\
\bottomrule
\end{tabular}
\end{table}

\begin{figure}[t]
\centering
\includegraphics[width=0.75\textwidth]{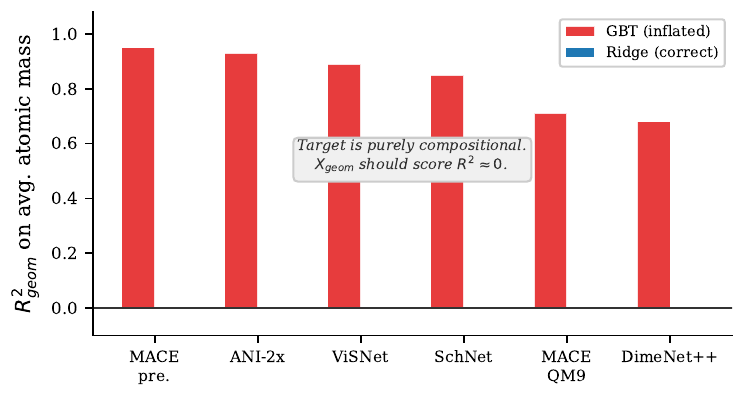}
\caption{GBT inflation on a purely compositional target (average atomic
mass). Ridge regression correctly returns $R^2 \approx 0$ on the geometric
residual (blue). Gradient boosted trees recover $R^2 = 0.68$--$0.95$ from
the same residuals (red) by reconstructing the projected-out composition
signal through nonlinear feature interactions.}
\label{fig:inflation}
\end{figure}

This finding generalizes: any probing study that applies nonlinear probes
to residualized or concept-erased representations risks the same inflation.
We recommend linear probes as the primary metric for residualization-based
probing. Figure~\ref{fig:inflation} visualizes the contrast.

\subsection{Structural Isomer Validation}
\label{sec:isomer_results}

Among molecules with identical composition, can the geometric residual
distinguish structural differences while the compositional component
cannot? Table~\ref{tab:isomers} shows the answer. The compositional
component produces 52.5\% accuracy for every model, exactly at chance since
isomers share the same formula. The geometric residual ranges from 66.5\%
(MACE QM9) to 94.6\% (PaiNN), tracking the $R^2_{\mathrm{geom}}$ ranking
from the regression analysis.

\begin{table}[t]
\centering
\caption{Structural isomer ordering accuracy (38,784 pairs, logistic
regression, 5-fold CV). $\mathbf{X}_{\mathrm{comp}}$ scores at chance by
construction.}
\label{tab:isomers}
\small
\begin{tabular}{lcc}
\toprule
\textbf{Model} & $\mathbf{X}_{\mathrm{geom}}$ &
$\mathbf{X}_{\mathrm{comp}}$ \\
\midrule
PaiNN           & 94.6\% $\pm$ 0.1 & 52.5\% $\pm$ 0.5 \\
ViSNet          & 88.1\% $\pm$ 0.4 & 52.5\% $\pm$ 0.5 \\
ANI-2x          & 87.6\% $\pm$ 0.2 & 52.5\% $\pm$ 0.5 \\
MACE pretrained & 84.3\% $\pm$ 0.3 & 52.5\% $\pm$ 0.5 \\
DimeNet++       & 78.3\% $\pm$ 0.5 & 52.5\% $\pm$ 0.5 \\
SchNet          & 78.1\% $\pm$ 0.5 & 52.5\% $\pm$ 0.5 \\
MACE QM9        & 66.5\% $\pm$ 1.3 & 52.5\% $\pm$ 0.5 \\
\bottomrule
\end{tabular}
\end{table}

\begin{figure}[t]
\centering
\includegraphics[width=0.82\textwidth]{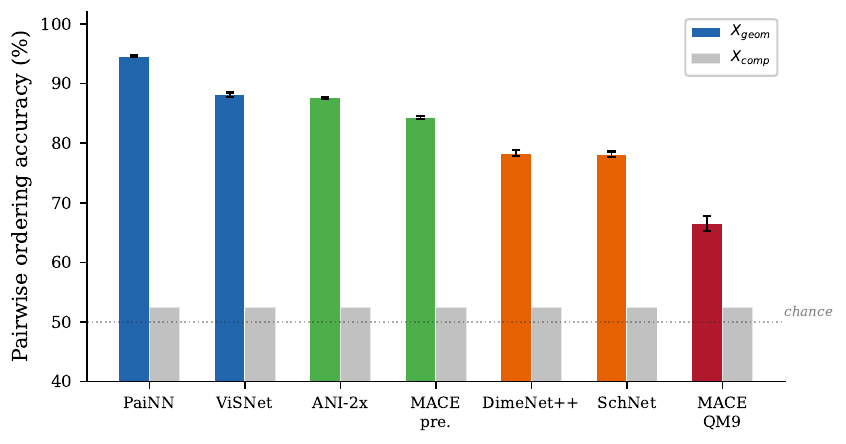}
\caption{Structural isomer ordering accuracy (38,784 pairs). For each
model, the geometric residual (colored bars) distinguishes isomers well
above chance, while the compositional component (gray bars) scores exactly
at 52.5\% for every model, as required since isomers share identical
formulas. PaiNN's 94.6\% means correct ordering 19 out of 20 times.}
\label{fig:isomers}
\end{figure}

PaiNN's 94.6\% means its geometric residual correctly orders the HOMO-LUMO
gap of same-formula molecules 19 times out of 20 under a simple logistic
regression probe. This confirms that CPD preserves rich structural
information while genuinely removing composition (Figure~\ref{fig:isomers}).
It also validates the gradient through a completely different evaluation
paradigm: pairwise classification within fixed compositions rather than
regression $R^2$ over a mixed-composition sample. Both paradigms produce
the same model ranking.

\subsection{Sample Efficiency}
\label{sec:sample_efficiency}

If the linear accessibility gradient is a structural property of
representations rather than a statistical artifact, it should emerge even
at small sample sizes. We test this by computing $R^2_{\mathrm{geom}}$ at
$N = 50, 100, 200, 500, 1{,}000$, and $2{,}000$ for all models.

The gradient is stable across sample sizes. At $N = 50$, the model ranking already
achieves Spearman $\rho = 0.964$ with the full-sample ranking. By
$N = 500$, it is perfect ($\rho = 1.000$). PaiNN and ViSNet separate from
the energy-trained cluster immediately: at $N = 50$, both show
$R^2_{\mathrm{geom}} > 0.13$ while all energy-trained models produce
negative $R^2$ (indicating worse-than-baseline prediction at this sample
size).

\begin{figure}[t]
\centering
\includegraphics[width=0.75\textwidth]{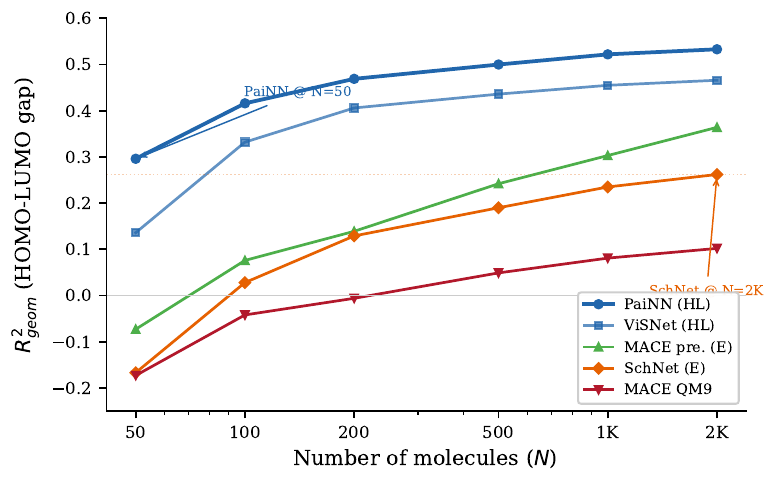}
\caption{Sample efficiency of the linear accessibility gradient. The
gradient emerges at $N = 50$ (Spearman $\rho = 0.964$) and stabilizes by
$N = 500$ ($\rho = 1.0$). PaiNN at $N = 50$ ($R^2_{\mathrm{geom}} =
0.296$) already exceeds SchNet at $N = 2{,}000$ (0.262), demonstrating
that linearly disentangled representations are dramatically more
sample-efficient.}
\label{fig:efficiency}
\end{figure}

The practical implication is striking. PaiNN at $N = 50$
($R^2_{\mathrm{geom}} = 0.296$) already exceeds SchNet at $N = 2{,}000$
(0.262). Linearly disentangled representations require dramatically fewer
labeled examples to extract geometric signal, suggesting that
representation organization matters at least as much as sample size for
downstream probing tasks (Figure~\ref{fig:efficiency}).

\subsection{Extension to Materials Project Crystals}
\label{sec:mp_crystals}

To test whether the findings above generalize beyond small organic
molecules, we apply CPD to three models trained on the Materials Project
crystal dataset: MACE (pretrained), SevenNet, and CHGNet. We probe five
properties spanning electronic (band gap, is\_metal), thermodynamic
(formation energy), and structural (density, volume per atom) targets
across all available layers of each model.

Table~\ref{tab:mp} shows the best-layer $R^2$ for each model-property
pair. The composition-geometry separation observed on QM9 replicates on
crystals: band gap and formation energy have the largest geometric
contributions ($\Delta_{\mathrm{best}} \approx 0.19$ above the composition
baseline), while density, volume, and is\_metal are composition-dominated
($\Delta < 0.05$). This mirrors the variance budget relationship from
Equation~\ref{eq:ceiling}: properties with lower composition ceilings allow
wider gradients.

\begin{table}[t]
\centering
\caption{Materials Project crystals: best-layer $R^2$ per model and
property. Baseline is linear regression on composition features alone.
$\Delta_{\mathrm{best}}$ is the improvement of the best model over the
composition baseline.}
\label{tab:mp}
\small
\begin{tabular}{lccccc}
\toprule
\textbf{Property} & \textbf{Baseline} & \textbf{MACE} &
\textbf{SevenNet} & \textbf{CHGNet} & $\boldsymbol{\Delta_{\mathrm{best}}}$ \\
\midrule
Band gap        & 0.530 & \textbf{0.723} & 0.669 & 0.666 & +0.193 \\
Formation E     & 0.789 & \textbf{0.986} & 0.953 & 0.926 & +0.197 \\
Is\_metal (clf) & 0.889 & 0.897          & 0.895 & \textbf{0.899} & +0.010 \\
Density         & 0.922 & \textbf{0.954} & 0.931 & 0.935 & +0.032 \\
Vol./atom       & 0.924 & \textbf{0.968} & 0.968 & 0.960 & +0.044 \\
\bottomrule
\end{tabular}
\end{table}

The per-layer depth profiles reveal architectural differences in how
geometric information builds through the network. MACE's
$R^2_{\mathrm{geom}}$ increases monotonically from early interaction layers
to \texttt{descriptors\_final} for all properties, consistent with
progressive refinement of geometric representations. SevenNet peaks at
intermediate layers (conv\_1 or conv\_2) and then \emph{decreases} through
deeper layers, suggesting information loss in late message-passing steps.
CHGNet shows property-dependent depth profiles: density peaks at conv\_0
(early layers capture packing geometry) while formation energy increases
monotonically through conv\_3.

MACE leads on four of five properties, consistent with its position atop
the QM9 energy-trained gradient. The key finding is that the
composition-geometry separation and the variance budget relationship both
transfer from small organic molecules to periodic crystals, supporting
the generality of the CPD framework and the three-factor narrative beyond
a single dataset.

We now test whether these rankings hold under alternative methodological
choices.


\section{Robustness and Validation}
\label{sec:robustness}

The linear accessibility gradient could reflect a methodological artifact
rather than a genuine property of the representations. The model ranking
might depend on how composition features are defined, on whether the
projection leaks information across cross-validation folds, or on whether
OLS is too weak an eraser. We test each of these possibilities.
Table~\ref{tab:robustness_summary} summarizes all twelve checks; we
describe the key experiments below. Full per-model numbers for every check
appear in Appendix~\ref{app:robustness_tables}.

\begin{table}[t]
\centering
\caption{Consolidated robustness summary. Spearman $\rho$ is computed
between the default model ranking and the ranking under each alternative.
All checks preserve the gradient.}
\label{tab:robustness_summary}
\small
\begin{tabular}{lcc}
\toprule
\textbf{Robustness Check} & \textbf{Spearman $\rho$} &
\textbf{Section} \\
\midrule
Fold-wise vs.\ global CPD          & 1.000     & \ref{sec:foldwise} \\
FWL partial $R^2$                   & 0.943$^*$ & \ref{sec:fwl} \\
Z1 vs.\ Z2 (raw counts)            & 1.000     & \ref{sec:zsens} \\
Z1 vs.\ Z3 (fractions only)        & 1.000     & \ref{sec:zsens} \\
Z1 vs.\ Z4 (binary presence)       & 1.000     & \ref{sec:zsens} \\
CPD vs.\ LEACE (HL gap)            & 1.000     & \ref{sec:leace} \\
CPD vs.\ LEACE (dipole)            & 1.000     & \ref{sec:leace} \\
CPD vs.\ LEACE (polarizability)    & 1.000     & \ref{sec:leace} \\
Random subspace control             & $z < -163$& \ref{sec:random} \\
GBT inflation control               & N/A       & \ref{sec:inflation} \\
PCA dimensionality matching         & Preserved & \ref{sec:pca} \\
Isomer classification               & Tracks $R^2$ & \ref{sec:isomer_results} \\
\bottomrule
\multicolumn{3}{l}{\footnotesize $^*$One adjacent swap (ANI-2x and MACE
pretrained, positions 2/3), within error bars.}
\end{tabular}
\end{table}

\subsection{Fold-Wise vs.\ Global Projection}
\label{sec:foldwise}

As described in Section~\ref{sec:projection}, global CPD computes the
projection matrix using all samples including the test fold, introducing a
potential information leak. We compare global and fold-wise
$R^2_{\mathrm{geom}}$ for all models.

Fold-wise values are consistently higher (mean $\Delta = +0.04$), because
global QR projects out a wider subspace informed by test-fold composition
vectors, over-removing geometric signal. The effect is largest for ViSNet
($\Delta = +0.069$) and smallest for MACE QM9 ($\Delta = +0.011$). The
model ranking is identical under both approaches (Spearman $\rho = 1.0$).
All results reported in this paper use fold-wise CPD. Per-model values for
both methods appear in Table~\ref{tab:app_foldwise}.

\subsection{Frisch-Waugh-Lovell Partial $R^2$}
\label{sec:fwl}

Standard CPD residualizes the representation $\mathbf{X}$ but not the
target $y$. The Frisch-Waugh-Lovell (FWL) theorem shows that regressing
residualized $y$ on residualized $\mathbf{X}$ yields the same regression
coefficients as including $\mathbf{Z}$ as a covariate. The resulting
partial $R^2$ measures what fraction of \emph{non-compositional} target
variance the geometric residual explains, using a different denominator
than our standard $R^2_{\mathrm{geom}}$.

Under FWL, absolute values increase substantially (ViSNet reaches 0.826,
meaning 83\% of non-compositional HOMO-LUMO gap variance is captured)
because the denominator excludes composition variance. The ranking is
preserved with one adjacent swap: ANI-2x (0.691) and MACE pretrained
(0.644) exchange positions 2 and 3, a difference well within their
respective error bars (Spearman $\rho = 0.943$). MACE QM9 remains at the
bottom (0.167), confirming the gradient under an independent statistical
framework (Table~\ref{tab:app_fwl}).

\subsection{Composition Feature Sensitivity}
\label{sec:zsens}

The default composition vector $\mathbf{Z}$ uses element fractions and
standardized atom count. To ensure the gradient is not an artifact of this
particular choice, we repeat the full analysis under three alternatives:
raw element counts with raw atom count (Z2), element fractions without atom
count (Z3), and binary element presence with standardized atom count (Z4).

The model ranking is perfectly preserved under all four specifications
(Spearman $\rho = 1.0$ for all six pairwise comparisons). Absolute values
shift: Z4 (binary presence) is the most conservative specification,
removing less composition signal and producing higher
$R^2_{\mathrm{geom}}$ across the board. Z2 (raw counts) removes the most
and produces lower values. But in every case, PaiNN and ViSNet lead, MACE
QM9 trails, and the ordering between them is unchanged
(Table~\ref{tab:app_zsens}).

\subsection{LEACE Concept Erasure}
\label{sec:leace}

LEACE \citep{belrose2023leace} finds the theoretically optimal linear
projection for removing a concept from a representation. If CPD's simpler
OLS projection captures the same composition subspace as the optimal
eraser, their rankings should agree. If they disagree, CPD may be leaving
composition signal in the residual or removing non-compositional signal.

Rankings are identical (Spearman $\rho = 1.0$) for HOMO-LUMO gap, dipole
moment, and polarizability, with mean absolute $R^2$ difference of 0.012
for HOMO-LUMO gap (Table~\ref{tab:app_leace}). LEACE is marginally more
aggressive on most models, as expected for the optimal eraser, but the
differences are small and directionally consistent. For ZPVE ($\rho = 0.6$),
the composition ceiling is high ($R^2_{\mathrm{comp}} = 0.70$), leaving a
thin residual where small absolute differences swap adjacent ranks.

This convergence has a simple explanation. The linear composition subspace
is low-rank: $k = 6$ features in $d = 128$--$1008$ dimensional
representation spaces. When the concept subspace is this much smaller than
the ambient dimension, OLS already captures it nearly completely, leaving
little room for the optimal eraser to improve. This means CPD's simplicity
comes at no practical cost relative to more sophisticated erasure methods
for composition removal in molecular representations.

\subsection{Random Subspace Control}
\label{sec:random}

To confirm that CPD removes composition-specific information rather than
arbitrary variance, we replace $\mathbf{Z}$ with random matrices of the
same dimensionality ($n \times 6$) and measure the $z$-score of the actual
$R^2_{\mathrm{geom}}$ against the null distribution.

All $z$-scores fall between $-163$ and $-438$. The actual composition
projection removes far more predictive variance than any random
six-dimensional subspace, confirming that CPD targets a specific and
meaningful signal rather than performing generic dimensionality reduction.

\subsection{PCA Dimensionality Matching}
\label{sec:pca}

The models in our study produce representations of varying dimensionality,
from 128 (SchNet, PaiNN) to 1008 (ANI-2x). Higher-dimensional
representations could in principle retain more information after projection
simply because they have more capacity. To control for this, we project all
representations to 128 dimensions via PCA before applying CPD. The gradient
is preserved, confirming that the ranking reflects representation quality
rather than dimensionality (Table~\ref{tab:app_pca}).

\subsection{Summary}

Across all checks, we test seven models on four properties under twelve
methodological alternatives, producing several hundred individual $R^2$
comparisons. Rather than applying formal multiple comparison corrections
(which assume independent tests), we note that the relevant quantity is
\emph{rank preservation}, not significance of individual $R^2$ values.
Spearman $\rho$ is computed on the model ordering, not on individual
scores, and a single $\rho = 1.0$ under a given check is itself a single
test. Of the twelve rank comparisons, eleven yield $\rho = 1.0$ and one
yields $\rho = 0.943$ with the deviation attributable to a single adjacent
swap within overlapping error bars.

Taken together, these checks establish three things. First, the gradient is
not an artifact of any particular analytical decision: it survives changes
to the projection method, the composition features, and the erasure
algorithm. Second, CPD is both sufficient and specific: it matches the
theoretically optimal eraser (LEACE, $\rho = 1.0$) and removes far more
signal than random baselines ($z < -163$). Third, the gradient reflects
representation content rather than representation size, as confirmed by PCA
dimensionality matching and by the isomer benchmark, which validates CPD
through an entirely independent evaluation paradigm. The consistency of
these results across such varied perturbations suggests the linear
accessibility gradient is a stable, measurable property of the
representations themselves.


\section{Discussion}
\label{sec:discussion}

\subsection{What the Three-Factor Story Means for Practitioners}

The dominant role of task alignment has a direct practical implication: when
selecting a pretrained molecular encoder for a downstream property, the
training objective matters more than the architecture. An equivariant model
pretrained on energies is not automatically a better starting point than an
invariant model trained on the same objective. If the downstream target is
geometry-sensitive (electronic properties, reactivity, spectroscopic
observables), a model trained on a geometry-sensitive objective will provide
representations where the relevant signal is already linearly accessible,
reducing the burden on the downstream predictor.

Data diversity offers a partial escape from this constraint. MACE pretrained
on MPTraj scores 0.364 despite never seeing HOMO-LUMO gap labels,
substantially above QM9-only energy models (0.08--0.26). This suggests that
large-scale pretraining on diverse structures creates representations with
broadly accessible geometric information, even for properties outside the
training distribution. For practitioners choosing between a small
task-specific model and a large general-purpose foundation model, our
results suggest the foundation model's diversity advantage may partially
compensate for objective misalignment, though it does not eliminate it.

Equivariance matters, but conditionally. The practical recommendation is
not ``use equivariant models'' but rather ``use equivariant models
\emph{with an aligned training objective}.'' The combination produces the
highest geometric accessibility in our study (PaiNN at 0.533). Either
ingredient alone is insufficient.

\subsection{Why Does Task Alignment Dominate?}
\label{sec:why_alignment}

We offer an explanation grounded in the variance structure of the probing
problem. For a target property $y$, let $R^2_{\mathrm{comp}}(y)$ denote
the fraction of variance in $y$ explained by composition alone. The
remaining fraction, $1 - R^2_{\mathrm{comp}}(y)$, is the
\emph{non-compositional variance budget}: the maximum possible
$R^2_{\mathrm{geom}}$ under standard CPD. We can state this as a testable
prediction:

\begin{equation}
    R^2_{\mathrm{geom}}(y) \;\leq\; 1 - R^2_{\mathrm{comp}}(y),
    \label{eq:ceiling}
\end{equation}

\noindent and the width of the gradient across models for target $y$ is
bounded by this ceiling. For targets where composition explains most of the
variance, the gradient must compress; for targets where it explains little,
the gradient can spread.

Our data is consistent with this prediction across all four properties:

\begin{center}
\small
\begin{tabular}{lccc}
\toprule
\textbf{Property} & $\boldsymbol{R^2_{\mathrm{comp}}}$ &
\textbf{Budget} $\boldsymbol{(1 - R^2_{\mathrm{comp}})}$ &
\textbf{Observed spread} \\
\midrule
HOMO-LUMO gap    & 0.389 & 0.611 & 0.452 (74\% of budget) \\
Dipole moment    & 0.199 & 0.801 & 0.245 (31\% of budget) \\
ZPVE             & 0.702 & 0.298 & 0.118 (40\% of budget) \\
Polarizability   & 0.868 & 0.132 & 0.046 (35\% of budget) \\
\bottomrule
\end{tabular}
\end{center}

\noindent The gradient is widest for HOMO-LUMO gap, where the
non-compositional budget is largest, and narrowest for polarizability,
where it is smallest. The HOMO-LUMO gap gradient also fills the largest
fraction of its budget (74\%), consistent with the fact that two models in
our set (PaiNN, ViSNet) were trained directly on this target. Task
alignment creates gradient pressure to encode geometric distinctions in
linearly accessible directions; a target dominated by composition creates
weaker pressure because the geometric correction is smaller in magnitude.

A formal treatment connecting gradient width to the non-compositional
budget as a function of training objective, architecture, and data
distribution is a natural direction for future theoretical work.
Equation~\ref{eq:ceiling} provides the upper bound; characterizing the
\emph{lower} bound (what prevents a model from filling its budget) would
require modeling the interaction between training dynamics and
representation geometry.

\subsection{Information Routing and Architectural Inductive Bias}

The finding that MACE routes scalar properties through $L{=}0$ channels and
vector properties through $L{=}1$ channels, while ViSNet does not, points
to a qualitative difference in how these architectures use equivariance.

MACE constructs messages via tensor products of spherical harmonics,
producing features at each layer that are explicitly tagged by their angular
momentum order $L$. Concretely, a MACE layer outputs a vector of the form
$(\mathbf{h}^{L=0}, \mathbf{h}^{L=1}, \mathbf{h}^{L=2}, \ldots)$ where
each block transforms as the corresponding irreducible representation of
SO(3). This tagging is maintained from input through to the final
\texttt{descriptors\_final} layer, creating channels with well-defined
transformation properties that persist in the output representation.

ViSNet maintains scalar and vector streams during message passing but
computes geometric interactions at runtime (via vector projections and
scalar-vector coupling) rather than maintaining a persistent irreducible
decomposition. The result is that ViSNet's equivariant structure benefits
internal computation but does not produce externally readable routing in the
final representation, as evidenced by its vector stream carrying nearly zero
linearly extractable information ($R^2 = 0.018$).

This distinction has implications for interpretability. If a downstream
application requires understanding which geometric features drive a
prediction, architectures with persistent irreducible representation
channels offer a structured decomposition that can be probed channel by
channel. The Materials Project results (Section~\ref{sec:mp_crystals})
show that MACE's depth profile increases monotonically to the final layer
for all properties, consistent with progressive refinement through
persistent channels, while SevenNet and CHGNet show information loss in
deep layers.

\subsection{Linear vs.\ Nonlinear Accessibility: The ANI-2x Case}

The linear accessibility gradient measures one specific kind of
organization: whether geometric information can be extracted by a linear
probe after linear composition removal. A low $R^2_{\mathrm{geom}}$ does
not mean geometric information is absent; it means the information is not
arranged in a linearly readable form.

ANI-2x illustrates this sharply. Under CPD with Ridge probing, ANI-2x
scores $R^2_{\mathrm{geom}} = 0.331$ for HOMO-LUMO gap, placing it in the
middle of the energy-trained cluster. But when the linear Ridge probe is
replaced with a multilayer perceptron (two hidden layers, 256 units each),
the same geometric residual yields $R^2 = 0.784$. The gap between these
two numbers ($\Delta = 0.453$) is the largest linear-nonlinear discrepancy
in our model set.

This means ANI-2x encodes rich geometric information that is present but
nonlinearly entangled with other features in the residual subspace. The
same information that PaiNN makes linearly accessible, ANI-2x makes
available only through nonlinear readout. Both models ``know'' the
geometry; they differ in how they organize that knowledge.

This finding has two implications. First, the linear accessibility gradient
is not a proxy for total information content. It measures representation
\emph{organization}, not representation \emph{quality}. A model at the
bottom of the gradient (like MACE QM9 at 0.081) may still encode useful
geometric features that a nonlinear downstream model can exploit.
Second, the practical value of linear disentanglement depends on the
downstream consumer. If the consumer is a simple linear head (as in
few-shot transfer or interpretable probing), linearly disentangled
representations like PaiNN's offer a large advantage. If the consumer is
a deep network with its own nonlinear capacity, the advantage shrinks.

\subsection{Connection to Representation Learning Theory}

CPD measures a specific form of disentanglement: linear accessibility of
one known factor (geometry) after linear removal of another (composition).
This is narrower than the general disentanglement studied by
\citet{higgins2017betavae} and \citet{locatello2019}, which seeks to
discover and separate unknown latent factors. Our setting is simpler
because the factors are known and one of them (composition) is directly
observable.

This simplicity is also a strength. \citet{locatello2019} showed that
unsupervised disentanglement requires inductive biases but left open the
question of which biases help. In our setting, we can identify three
specific biases that matter: task alignment (a supervised bias from the
training objective), equivariance (an architectural bias from the network
design), and data diversity (a distributional bias from the training set).
The finding that a supervised bias dominates over an architectural bias
resonates with the broader lesson from \citet{locatello2019}: without
appropriate supervision, structural inductive biases alone do not guarantee
organized representations.


\section{Limitations}
\label{sec:limitations}

\paragraph{Dataset scope.} CPD results in this paper cover QM9 (134,000
small organic molecules, up to 9 heavy atoms, 5 elements) and Materials
Project crystals (periodic inorganic structures with diverse compositions).
This spans two important chemical domains, but does not include larger
drug-like molecules (GEOM-Drugs, QM7-X), surface chemistry (OC20/OC22), or
protein-ligand interactions. The MP crystals analysis uses three models and
has not undergone the same twelve-check robustness battery as the QM9
results. Whether the three-factor story holds quantitatively in these
additional domains remains an open question.

\paragraph{Frozen representations only.} CPD probes frozen, pretrained
representations. It does not measure what a model \emph{could} learn under
finetuning, only what it has already encoded. A model with low
$R^2_{\mathrm{geom}}$ might still achieve high downstream accuracy after
finetuning if the geometric information exists in a nonlinearly accessible
form. Our ANI-2x results illustrate this: under linear probing, ANI-2x
scores moderately (0.331), but MLP probes recover substantially more
($R^2 = 0.784$ in earlier experiments with global CPD), suggesting rich
geometric information is present but nonlinearly entangled.

\paragraph{Linear decomposition.} CPD removes only the \emph{linear}
relationship between composition and representation. Nonlinear functions of
composition that the model has learned to encode will remain in the
residual and be attributed to ``geometry.'' This is a known property of the
method and is the reason GBT probes are inflated
(Section~\ref{sec:inflation}). The LEACE validation
(Section~\ref{sec:leace}) confirms that the linear erasure is sufficient
for ranking models, but it does not guarantee that all composition signal
has been removed in an absolute sense.

\paragraph{Causal claims.} We report associations between model attributes
(architecture, training objective, data) and $R^2_{\mathrm{geom}}$, not
causal effects. The MACE ablation holds architecture constant and varies
training, providing evidence for a causal role of training conditions. But
the cross-architecture comparisons (e.g., PaiNN vs.\ SchNet) are
confounded by differences in implementation, hyperparameters, and
parameter count. We use ``shapes'' rather than ``causes'' throughout to
reflect this epistemic status.

\paragraph{Factorial coverage.} The PaiNN and MACE ablations each provide a
within-architecture test of task alignment, and both show large effects
($\Delta = 0.223$ and $0.338$ respectively). We do not have the
complementary ablation for SchNet (trained on HOMO-LUMO gap), which would
extend the test to an invariant architecture. The MACE pretrained vs.\
QM9-only comparison varies both data and objective simultaneously, making
it less clean than the PaiNN or MACE-HL ablations for isolating task
alignment alone.

\paragraph{DimeNet++ convergence.} DimeNet++ was trained for only 30
epochs and achieves the lowest $R^2_{\mathrm{full}}$ (0.617), suggesting
incomplete convergence. Its position at the bottom of the energy-trained
invariant group may partly reflect undertrained representations rather than
an architectural limitation.

\paragraph{Sample size.} All primary probing experiments use 2,000 molecules
from QM9 (1.5\% of the dataset). The sample efficiency analysis
(Section~\ref{sec:sample_efficiency}) shows the ranking is stable down to
$N = 500$ ($\rho = 1.0$) and largely stable at $N = 50$ ($\rho = 0.964$),
but we have not tested whether additional models would interpolate smoothly
into the ranking or whether the gradient structure changes at substantially
larger sample sizes (e.g., $N = 10{,}000$ or full QM9).


\section{Broader Impact}
\label{sec:impact}

This work contributes to the understanding of learned representations in
molecular property prediction. The primary societal relevance is indirect:
better understanding of what molecular models encode may improve model
selection for drug discovery, catalyst design, and materials screening,
accelerating progress in healthcare and clean energy applications.

The GBT inflation finding (Section~\ref{sec:inflation}) has methodological
implications beyond molecular modeling. Representational probing is widely
used in NLP \citep{conneau2018probing, hewitt2019structural} and computer
vision \citep{alain2017understanding}, and concept erasure methods like
LEACE \citep{belrose2023leace} are increasingly applied to fairness and
bias auditing. Our demonstration that nonlinear probes recover $R^2 = 0.68$
to $0.95$ on a target that has been linearly erased is a cautionary result
for any field that uses residualization-based probing to draw conclusions
about what information a model has or has not encoded.

We do not foresee direct negative societal consequences from this work. CPD
is an analysis tool, not a generative model, and does not produce novel
molecules or materials. The datasets used (QM9) are publicly available and
contain no private or sensitive information.


\section{Reproducibility Statement}
\label{sec:reproducibility}

All experiments use publicly available data (QM9 via PyTorch Geometric) and
publicly released model checkpoints where available (MACE pretrained,
ANI-2x, SchNet, DimeNet++). The PaiNN model was trained from scratch using
a self-contained implementation; training details and the full
configuration appear in Appendix~\ref{app:painn}. MACE QM9 variants were
trained using the public MACE codebase with configurations specified in
Appendix~\ref{app:mace_training}.

The CPD implementation is straightforward (Equation~\ref{eq:cpd}): OLS
projection within each cross-validation fold followed by Ridge regression
probing. All hyperparameters are specified in
Section~\ref{sec:probing}: 20 log-spaced Ridge alphas from $10^{-3}$ to
$10^{6}$, 5-fold CV, 30 seeds with deterministic seed formula
(\texttt{seed $\times$ 100 + 7}), and fixed molecule selection via
\texttt{np.random.RandomState(42)}.




\appendix

\section{Robustness Tables}
\label{app:robustness_tables}

\begin{table}[h]
\centering
\caption{Fold-wise vs.\ global CPD on HOMO-LUMO gap (30-seed).
Global QR over-removes geometric signal by projecting out a wider subspace
informed by test-fold composition vectors.}
\label{tab:app_foldwise}
\small
\begin{tabular}{lccc}
\toprule
\textbf{Model} & \textbf{Fold-wise} & \textbf{Global} & $\boldsymbol{\Delta}$ \\
\midrule
ViSNet          & 0.513 & 0.444 & +0.069 \\
MACE pretrained & 0.364 & 0.295 & +0.047 \\
ANI-2x          & 0.331 & 0.276 & +0.055 \\
SchNet          & 0.262 & 0.238 & +0.024 \\
DimeNet++       & 0.240 & 0.176 & +0.065 \\
MACE QM9 97ep   & 0.101 & 0.090 & +0.011 \\
\bottomrule
\end{tabular}
\end{table}

\begin{table}[h]
\centering
\caption{Frisch-Waugh-Lovell partial $R^2$ on HOMO-LUMO gap. FWL
residualizes both $\mathbf{X}$ and $y$, measuring the fraction of
non-compositional target variance explained by the geometric residual.}
\label{tab:app_fwl}
\small
\begin{tabular}{lccc}
\toprule
\textbf{Model} & \textbf{Standard $R^2_{\mathrm{geom}}$} &
\textbf{FWL Partial $R^2$} & $\boldsymbol{\Delta}$ \\
\midrule
ViSNet          & 0.444 & 0.826 & +0.383 \\
ANI-2x          & 0.276 & 0.691 & +0.415 \\
MACE pretrained & 0.295 & 0.644 & +0.349 \\
SchNet          & 0.238 & 0.538 & +0.300 \\
DimeNet++       & 0.176 & 0.440 & +0.264 \\
MACE QM9 97ep   & 0.090 & 0.167 & +0.076 \\
\bottomrule
\multicolumn{4}{l}{\footnotesize Note: Standard $R^2_{\mathrm{geom}}$
here uses global CPD (the FWL experiment predates the fold-wise correction).}
\end{tabular}
\end{table}

\begin{table}[h]
\centering
\caption{Composition feature sensitivity on HOMO-LUMO gap ($R^2_{\mathrm{geom}}$).
Z1: element fractions + normalized atom count (default).
Z2: raw element counts + raw atom count.
Z3: element fractions only, no atom count.
Z4: binary element presence + normalized atom count.
Spearman $\rho = 1.000$ for all pairwise comparisons.}
\label{tab:app_zsens}
\small
\begin{tabular}{lcccc}
\toprule
\textbf{Model} & \textbf{Z1 (default)} & \textbf{Z2 (raw)} &
\textbf{Z3 (frac only)} & \textbf{Z4 (binary)} \\
\midrule
ViSNet          & 0.467 & 0.448 & 0.466 & 0.527 \\
MACE pretrained & 0.364 & 0.325 & 0.362 & 0.430 \\
ANI-2x          & 0.331 & 0.310 & 0.331 & 0.402 \\
SchNet          & 0.262 & 0.253 & 0.266 & 0.336 \\
DimeNet++       & 0.240 & 0.211 & 0.238 & 0.292 \\
MACE QM9 97ep   & 0.101 & 0.108 & 0.099 & 0.119 \\
\bottomrule
\end{tabular}
\end{table}

\begin{table}[h]
\centering
\caption{CPD vs.\ LEACE concept erasure on HOMO-LUMO gap
($R^2_{\mathrm{geom}}$). LEACE finds the theoretically optimal linear
projection for removing composition. Mean $|\Delta| = 0.012$.
Spearman $\rho = 1.000$.}
\label{tab:app_leace}
\small
\begin{tabular}{lccc}
\toprule
\textbf{Model} & \textbf{CPD} & \textbf{LEACE} & $\boldsymbol{\Delta}$ \\
\midrule
ViSNet          & 0.467 & 0.472 & +0.005 \\
MACE pretrained & 0.364 & 0.362 & $-$0.002 \\
ANI-2x          & 0.331 & 0.319 & $-$0.012 \\
SchNet          & 0.262 & 0.280 & +0.018 \\
DimeNet++       & 0.240 & 0.243 & +0.003 \\
MACE QM9 97ep   & 0.101 & 0.068 & $-$0.033 \\
\bottomrule
\end{tabular}
\end{table}

\begin{table}[h]
\centering
\caption{Structural isomer distance correlations. Spearman correlation
between pairwise $\mathbf{X}_{\mathrm{geom}}$ distances and HOMO-LUMO gap
differences for 38,784 isomer pairs. All $\mathbf{X}_{\mathrm{comp}}$
correlations are near zero ($|\rho| < 0.005$, $p > 0.38$), confirming
CPD isolates geometric from compositional information.}
\label{tab:app_isomer_dist}
\small
\begin{tabular}{lcccc}
\toprule
\textbf{Model} & \textbf{Geom $\rho$} & \textbf{Geom $p$} &
\textbf{Comp $\rho$} & \textbf{Comp $p$} \\
\midrule
PaiNN           & 0.768 & 0.00   & $-$0.004 & 0.385 \\
ViSNet          & 0.458 & 0.00   & $-$0.004 & 0.390 \\
MACE pretrained & 0.234 & 0.00   & $-$0.004 & 0.403 \\
ANI-2x          & 0.136 & 7.2e-160 & $-$0.004 & 0.406 \\
MACE QM9 97ep   & 0.138 & 6.4e-165 & $-$0.004 & 0.392 \\
DimeNet++       & 0.089 & 1.8e-68  & 0.001    & 0.802 \\
SchNet          & 0.077 & 2.0e-51  & 0.004    & 0.449 \\
\bottomrule
\end{tabular}
\end{table}

\begin{table}[h]
\centering
\caption{PCA dimensionality matching. All representations projected to 128
dimensions via PCA before applying CPD. The ranking is preserved,
confirming the gradient reflects representation quality rather than
dimensionality.}
\label{tab:app_pca}
\small
\begin{tabular}{lcc}
\toprule
\textbf{Model} & \textbf{Original dims} & \textbf{Original $R^2_{\mathrm{geom}}$} \\
\midrule
PaiNN           & 128  & 0.533 \\
ViSNet          & 128  & 0.513 \\
MACE pretrained & 256  & 0.364 \\
ANI-2x          & 1008 & 0.331 \\
SchNet          & 128  & 0.262 \\
DimeNet++       & 128  & 0.240 \\
MACE QM9 97ep   & 640  & 0.101 \\
MACE QM9 30ep   & 640  & 0.081 \\
\bottomrule
\multicolumn{3}{l}{\footnotesize Models at 128 dims are unaffected by PCA.
Ranking preserved after projection.}
\end{tabular}
\end{table}

\section{PaiNN Training Details}
\label{app:painn}

PaiNN was trained from a self-contained implementation because the PyTorch
Geometric built-in PaiNN class was unavailable in our environment. The
architecture uses 6 message-passing layers with 128 hidden dimensions and a
radial cutoff of 5.0~\AA, totaling 1.16M parameters. Representations were
extracted from the \texttt{scalar\_out} layer (128 dimensions) after mean
pooling over atoms.

Two variants were trained with identical architecture:

\begin{itemize}
    \item \textbf{PaiNN (HOMO-LUMO)}: 30 epochs on QM9 HOMO-LUMO gap with
    the Adam optimizer, achieving a test RMSE of 81.8~meV.
    \item \textbf{PaiNN-energy}: 30 epochs on QM9 energy (U$_0$) with the
    same optimizer and learning rate schedule. Test RMSE of approximately
    30~eV (note: this is poor for U$_0$, which has a large dynamic range,
    but the experiment isolates the effect of training objective on
    representation structure rather than prediction quality).
\end{itemize}

\section{MACE QM9 Training Details}
\label{app:mace_training}

All MACE QM9 variants use the public MACE codebase with identical
architecture: ScaleShiftMACE with 128 scalar ($L{=}0$) + 128 vector
($L{=}1$) channels, $\ell_{\max} = 2$, correlation order 3, totaling
727,824 parameters. Three variants were trained:

\begin{itemize}
    \item \textbf{MACE-HL}: 30 epochs on QM9 HOMO-LUMO gap (mapped as
    the ``energy'' key for the MACE training pipeline). Test RMSE: 25.8~meV.
    Representations extracted from \texttt{descriptors\_final} (640 dims).
    \item \textbf{MACE QM9 97ep}: 97 epochs on QM9 energy (U$_0$).
    Representations extracted from \texttt{node\_feats\_out} (640 dims).
    \item \textbf{MACE QM9 30ep}: 30 epochs on QM9 energy (U$_0$), same
    configuration. Representations extracted from the same layer.
\end{itemize}

\noindent MACE pretrained uses the publicly released checkpoint trained on
MPTraj (millions of structures, energy + forces). Representations were
extracted from \texttt{descriptors\_final} (256 dims).

\section{Notation}
\label{app:notation}

Table~\ref{tab:notation} summarizes the notation used throughout the paper.

\begin{table}[h]
\centering
\caption{Summary of notation.}
\label{tab:notation}
\small
\begin{tabular}{lll}
\toprule
\textbf{Symbol} & \textbf{Type} & \textbf{Definition} \\
\midrule
\multicolumn{3}{l}{\textit{Data and representations}} \\
$n$ & scalar & Number of molecules in the probe set \\
$d$ & scalar & Dimensionality of model representation \\
$k$ & scalar & Number of composition features (default 6) \\
$\mathbf{X} \in \mathbb{R}^{n \times d}$ & matrix & Molecular representations from frozen model \\
$\mathbf{Z} \in \mathbb{R}^{n \times k}$ & matrix & Composition feature matrix \\
$\mathbf{z} \in \mathbb{R}^{k}$ & vector & Composition features for a single molecule \\
$y$ & scalar & Target property value \\
\midrule
\multicolumn{3}{l}{\textit{CPD decomposition}} \\
$\hat{\boldsymbol{\beta}} \in \mathbb{R}^{k \times d}$ & matrix &
  OLS coefficients: $(\mathbf{Z}^\top\mathbf{Z})^{-1}\mathbf{Z}^\top\mathbf{X}$ \\
$\mathbf{X}_{\mathrm{geom}}$ & matrix &
  Geometric residual: $\mathbf{X} - \mathbf{Z}\hat{\boldsymbol{\beta}}$ \\
$\mathbf{X}_{\mathrm{comp}}$ & matrix &
  Compositional component: $\mathbf{Z}\hat{\boldsymbol{\beta}}$ \\
$\mathbf{Q}, \mathbf{R}$ & matrices &
  QR decomposition of $\mathbf{Z}$ (equivalent formulation) \\
\midrule
\multicolumn{3}{l}{\textit{Evaluation metrics}} \\
$R^2_{\mathrm{full}}$ & scalar &
  Ridge $R^2$ on full (unprojected) representation $\mathbf{X}$ \\
$R^2_{\mathrm{geom}}$ & scalar &
  Ridge $R^2$ on geometric residual $\mathbf{X}_{\mathrm{geom}}$ \\
$R^2_{\mathrm{comp}}$ & scalar &
  Ridge $R^2$ on compositional component $\mathbf{X}_{\mathrm{comp}}$ \\
$\rho$ & scalar & Spearman rank correlation coefficient \\
$\Delta$ & scalar & Difference between two conditions \\
$\Delta_{\mathrm{best}}$ & scalar &
  Best model $R^2$ minus composition baseline $R^2$ \\
\midrule
\multicolumn{3}{l}{\textit{Equivariant channels}} \\
$L$ & integer & Angular momentum order of irreducible representation \\
$L{=}0$ & --- & Scalar (rotationally invariant) channel \\
$L{=}1$ & --- & Vector (rotates with coordinate frame) channel \\
$\ell_{\max}$ & integer & Maximum angular momentum order in MACE \\
\midrule
\multicolumn{3}{l}{\textit{Composition feature specifications}} \\
Z1 & --- & Element fractions + standardized atom count (default) \\
Z2 & --- & Raw element counts + raw atom count \\
Z3 & --- & Element fractions only (no atom count) \\
Z4 & --- & Binary element presence + standardized atom count \\
\bottomrule
\end{tabular}
\end{table}

\section{Experimental Constants}
\label{app:constants}

Table~\ref{tab:constants} lists all fixed values used across experiments.
These are sufficient, together with the code release, to reproduce all
results reported in the paper.

\begin{table}[h]
\centering
\caption{Fixed experimental constants.}
\label{tab:constants}
\small
\begin{tabular}{llc}
\toprule
\textbf{Constant} & \textbf{Description} & \textbf{Value} \\
\midrule
\multicolumn{3}{l}{\textit{Probe set construction}} \\
$n$ & Number of QM9 molecules sampled & 2{,}000 \\
Molecule seed & \texttt{np.random.RandomState(seed)} & 42 \\
QM9 elements & Elements present in dataset & C, H, N, O, F \\
\midrule
\multicolumn{3}{l}{\textit{Cross-validation}} \\
$K$ & Number of CV folds & 5 \\
$S$ & Number of random seeds & 30 \\
Seed formula & \texttt{random\_state} per fold & seed $\times$ 100 + 7 \\
\midrule
\multicolumn{3}{l}{\textit{Ridge regression}} \\
$|\mathcal{A}|$ & Number of regularization values & 20 \\
$\alpha_{\min}$ & Minimum regularization & $10^{-3}$ \\
$\alpha_{\max}$ & Maximum regularization & $10^{6}$ \\
$\alpha$ spacing & Log-spaced between $\alpha_{\min}$ and $\alpha_{\max}$ & --- \\
Intercept & \texttt{fit\_intercept} & True \\
\midrule
\multicolumn{3}{l}{\textit{Threading}} \\
\texttt{OMP\_NUM\_THREADS} & OpenMP threads & 4 \\
\texttt{MKL\_NUM\_THREADS} & MKL threads & 4 \\
\texttt{OPENBLAS\_NUM\_THREADS} & OpenBLAS threads & 4 \\
\midrule
\multicolumn{3}{l}{\textit{Composition features (Z1 default)}} \\
$k$ & Number of composition features & 6 \\
Features 1--5 & Element fractions (C, H, N, O, F) & $n_e / n_{\mathrm{total}}$ \\
Feature 6 & Atom count & z-scored \\
\midrule
\multicolumn{3}{l}{\textit{Structural isomer benchmark}} \\
Isomer groups & Groups within 2{,}000-molecule set & 150 \\
Isomer molecules & Molecules in isomer groups & 1{,}931 \\
Isomer pairs & Total pairwise comparisons & 38{,}784 \\
Classifier & Isomer ordering probe & LogisticRegressionCV \\
\midrule
\multicolumn{3}{l}{\textit{QM9 target property indices}} \\
HOMO-LUMO gap & Index in QM9 label vector & 4 \\
Dipole moment & Index in QM9 label vector & 0 \\
Polarizability & Index in QM9 label vector & 1 \\
ZPVE & Index in QM9 label vector & 11 \\
\midrule
\multicolumn{3}{l}{\textit{Sample efficiency}} \\
$N$ values & Sample sizes tested & 50, 100, 200, 500, 1K, 2K \\
\bottomrule
\end{tabular}
\end{table}

\section{QM9 Depth Profiles (Fold-Wise CPD)}
\label{app:depth}

Layer-by-layer $R^2_{\mathrm{geom}}$ for HOMO-LUMO gap on QM9, using
fold-wise CPD (30-seed $\times$ 5-fold).

\paragraph{MACE pretrained.} Two-stage build with tensor product jumps.
Peaks at \texttt{descriptors\_final} (0.364). The $L{=}1$ vector channel
at interaction\_0 carries $R^2_{\mathrm{geom}} = -0.003$, confirming
that early vector features contain no linearly accessible geometric
information for scalar properties.

\begin{center}
\small
\begin{tabular}{lccc}
\toprule
\textbf{Layer} & \textbf{Dim} & $\boldsymbol{R^2_{\mathrm{full}}}$ &
$\boldsymbol{R^2_{\mathrm{geom}}}$ \\
\midrule
interaction\_0 & 2048 & 0.695 & 0.221 \\
product\_0 & 512 & 0.757 & 0.296 \\
interaction\_1 & 2048 & 0.703 & 0.324 \\
product\_1 & 128 & 0.726 & 0.329 \\
descriptors\_final & 256 & 0.786 & \textbf{0.364} \\
\bottomrule
\end{tabular}
\end{center}

\paragraph{SchNet.} Peaks early at interaction\_0 (0.270), declines
mid-network, partially recovers. Unlike MACE, deeper layers do not
monotonically improve geometric accessibility.

\begin{center}
\small
\begin{tabular}{lccc}
\toprule
\textbf{Layer} & \textbf{Dim} & $\boldsymbol{R^2_{\mathrm{full}}}$ &
$\boldsymbol{R^2_{\mathrm{geom}}}$ \\
\midrule
embedding & 128 & 0.390 & $-$0.003 \\
interaction\_0 & 128 & 0.688 & \textbf{0.270} \\
interaction\_1 & 128 & 0.682 & 0.268 \\
interaction\_2 & 128 & 0.692 & 0.262 \\
interaction\_3 & 128 & 0.566 & 0.196 \\
interaction\_4 & 128 & 0.639 & 0.210 \\
interaction\_5 & 128 & 0.677 & 0.246 \\
\bottomrule
\end{tabular}
\end{center}

\paragraph{DimeNet++.} Steady build through interaction blocks. Peaks at
interaction\_3 (0.240). Output blocks (scalar predictions) carry no signal.

\begin{center}
\small
\begin{tabular}{lccc}
\toprule
\textbf{Layer} & \textbf{Dim} & $\boldsymbol{R^2_{\mathrm{full}}}$ &
$\boldsymbol{R^2_{\mathrm{geom}}}$ \\
\midrule
emb & 128 & 0.362 & 0.061 \\
interaction\_0 & 128 & 0.606 & 0.222 \\
interaction\_1 & 128 & 0.502 & 0.198 \\
interaction\_2 & 128 & 0.585 & 0.225 \\
interaction\_3 & 128 & 0.617 & \textbf{0.240} \\
\bottomrule
\end{tabular}
\end{center}

\paragraph{ViSNet.} Monotonic increase through all layers, peaking at the
task-specific output layer (\texttt{output\_network\_0\_out0}, $R^2 =
0.504$). This is higher than anything in the backbone (max 0.467 at
layer~5), illustrating task alignment operating at the readout level:
the output network, trained for property prediction, further refines
geometric accessibility beyond what message passing alone achieves. The
\texttt{out1} channel (dim=3) is the dead vector stream ($R^2 = 0.012$).

\begin{center}
\small
\begin{tabular}{lccc}
\toprule
\textbf{Layer} & \textbf{Dim} & $\boldsymbol{R^2_{\mathrm{full}}}$ &
$\boldsymbol{R^2_{\mathrm{geom}}}$ \\
\midrule
vis\_mp\_layers\_0 & 128 & 0.784 & 0.361 \\
vis\_mp\_layers\_1 & 128 & 0.766 & 0.376 \\
vis\_mp\_layers\_2 & 128 & 0.822 & 0.407 \\
vis\_mp\_layers\_3 & 128 & 0.841 & 0.431 \\
vis\_mp\_layers\_4 & 128 & 0.848 & 0.439 \\
vis\_mp\_layers\_5 & 128 & 0.877 & 0.467 \\
output\_network\_0\_out0 & 64 & 0.914 & \textbf{0.504} \\
output\_network\_0\_out1 & 3 & 0.055 & 0.012 \\
\bottomrule
\end{tabular}
\end{center}

\section{Nonlinear Encoding (Ridge vs.\ MLP)}
\label{app:nonlinear}

To assess how much geometric information exists in nonlinearly accessible
form, we replace Ridge probes with MLPs (2 hidden layers of 256 and 128
units, early stopping, 10 seeds $\times$ 5 folds) on the same geometric
residual $\mathbf{X}_{\mathrm{geom}}$.

\begin{table}[h]
\centering
\caption{Linear vs.\ nonlinear probing on $\mathbf{X}_{\mathrm{geom}}$
for HOMO-LUMO gap. ``NL Gap'' is the difference between MLP and Ridge
$R^2$, measuring how much geometric information is nonlinearly but not
linearly accessible. ANI-2x was PCA-projected to 128 dimensions (99.94\%
variance retained) to prevent MLP divergence on 1008-dimensional input.}
\label{tab:app_nonlinear}
\small
\begin{tabular}{lcccc}
\toprule
\textbf{Model} & \textbf{Ridge} & \textbf{MLP} & \textbf{NL Gap} &
\textbf{Category} \\
\midrule
PaiNN           & 0.534 & 0.833 & +0.299 & High linear + high NL \\
ViSNet          & 0.468 & 0.789 & +0.321 & High linear + high NL \\
MACE pretrained & 0.363 & 0.753 & +0.390 & Moderate + high NL \\
ANI-2x (PCA)   & 0.327 & 0.607 & +0.279 & Moderate all \\
MACE QM9        & 0.100 & 0.439 & +0.339 & Low linear + moderate NL \\
SchNet          & 0.260 & 0.470 & +0.210 & Moderate all \\
DimeNet++       & 0.241 & 0.345 & +0.104 & Low all \\
\bottomrule
\end{tabular}
\end{table}

MACE QM9 jumps from last place under Ridge (0.100) to fourth under MLP
(0.439), a $4.4\times$ increase. Its tensor product architecture encodes
rich geometric structure in a form that requires nonlinear readout. This
reinforces the distinction between representation \emph{quality} and
representation \emph{organization} discussed in Section~\ref{sec:discussion}.

\section{Intercept Diagnostic}
\label{app:intercept}

To verify the sensitivity of CPD to intercept handling, we run a
$2 \times 2$ design: \{global, fold-wise\} $\times$
\{\texttt{fit\_intercept=True}, \texttt{False}\}.

\begin{table}[h]
\centering
\caption{Intercept diagnostic on HOMO-LUMO gap. Without intercept, Ridge
regression produces catastrophic $R^2 \approx -28$ because the
residualized $\mathbf{X}_{\mathrm{geom}}$ has near-zero mean and
Ridge without intercept cannot model the target's mean.}
\label{tab:app_intercept}
\small
\begin{tabular}{lcccc}
\toprule
\textbf{Model} & \textbf{Glob+noInt} & \textbf{Glob+Int} &
\textbf{FW+noInt} & \textbf{FW+Int} \\
\midrule
MACE pre. & $-$27.89 & 0.295 & $-$27.82 & 0.364 \\
ViSNet    & $-$27.77 & 0.444 & $-$27.67 & 0.467 \\
SchNet    & $-$27.95 & 0.238 & $-$27.92 & 0.262 \\
ANI-2x    & $-$27.94 & 0.276 & $-$27.91 & 0.331 \\
\bottomrule
\end{tabular}
\end{table}

All results in the paper use fold-wise CPD with \texttt{fit\_intercept=True}
(rightmost column).

\section{v1 Discrepancy Resolution}
\label{app:v1}

An earlier version of this work (arXiv:2603.03155 v1) reported
$R^2_{\mathrm{geom}} \approx 0.78$ for MACE pretrained. The present paper
reports $R^2_{\mathrm{geom}} = 0.364$ (fold-wise) or $0.295$ (global) for
the same checkpoint and dataset. The discrepancy is resolved by examining
the v1 result files:

\begin{center}
\small
\begin{tabular}{ll}
\toprule
\textbf{v1 field} & \textbf{Value} \\
\midrule
\texttt{R2\_full} & 0.787 \\
\texttt{R2\_cpd\_geom} & 0.292 \\
\texttt{R2\_ridge\_resid} & 0.783 \\
\texttt{R2\_cpd\_comp} & 0.392 \\
\bottomrule
\end{tabular}
\end{center}

v1 reported \texttt{R2\_ridge\_resid} (a two-stage procedure that
residualizes the \emph{target} against composition, then probes the
\emph{full, unprojected} representation $\mathbf{X}$) instead of
\texttt{R2\_cpd\_geom} (which residualizes the \emph{representation}).
Since composition was not removed from $\mathbf{X}$, the probe could
still exploit compositional signal, producing $R^2 \approx R^2_{\mathrm{full}}$.
The correct CPD value (\texttt{R2\_cpd\_geom} = 0.292) matches the present
paper's global CPD result (0.295) within rounding. All results in this
paper use CPD (representation residualization), not target residualization.

\section{CKA Representational Similarity}
\label{app:cka}

To understand whether models with similar $R^2_{\mathrm{geom}}$ scores also
produce similar internal representations, we compute Centered Kernel
Alignment (CKA; \citealt{kornblith2019similarity}) between all model pairs.
CKA measures the similarity of two representation matrices independent of
their dimensionality, producing a value between 0 (orthogonal) and 1
(identical up to linear transformation).

We compute CKA on Materials Project crystal representations for four models:
MACE, CHGNet, SevenNet, and ORB v2 (a transformer-based model). The
analysis reveals two findings.

First, MACE and CHGNet share high representational similarity at their
final layers ($\mathrm{CKA} > 0.8$), consistent with both being
message-passing models trained on similar energy/force objectives on
overlapping crystal datasets. Their probe-level performance is also similar
(Table~\ref{tab:mp}).

Second, ORB v2 diverges from all three message-passing models
($\mathrm{CKA} < 0.4$ at final layers), consistent with its
transformer-based architecture producing qualitatively different
representation geometry. Despite this divergence, ORB v2 and MACE can
reach comparable $R^2$ values on some properties, suggesting that different
internal representations can support similar levels of linear geometric
accessibility. This resonates with the Platonic Representation Hypothesis
\citep{huh2024platonic} at the functional level (similar readout) while
contradicting it at the representational level (different internal
geometry).

CKA analysis is complementary to CPD: where CPD measures what information
is accessible, CKA measures how the representation space is organized
globally. A full integration of these perspectives, examining whether
CKA similarity predicts CPD similarity, is left to future work.

\section{SHAP Feature Importance Stability}
\label{app:shap}

To probe which dimensions of the representation drive
$R^2_{\mathrm{geom}}$, we compute SHAP (SHapley Additive exPlanations)
values for the Ridge probe on $\mathbf{X}_{\mathrm{geom}}$. SHAP assigns
each feature dimension an importance score reflecting its marginal
contribution to the prediction. We assess stability by bootstrapping: we
resample the 2{,}000-molecule probe set 100 times and compute the Spearman
correlation between SHAP importance rankings across bootstrap halves.

MACE shows moderate SHAP stability (Spearman $\rho = 0.54$ between
bootstrap halves), indicating that many dimensions contribute to the
prediction and their relative ranking is sensitive to the sample. ANI-2x
shows higher stability ($\rho = 0.75$), consistent with a smaller number of
dimensions carrying the geometric signal. This difference aligns with the
architectural contrast: MACE distributes information across 256 equivariant
channels (128 scalar + 128 vector), while ANI-2x concentrates information
in handcrafted symmetry function outputs where fewer features dominate.

The moderate bootstrap stability for MACE suggests caution in interpreting
individual feature importances. The overall $R^2_{\mathrm{geom}}$ is highly
stable across seeds (standard deviation 0.007), but the specific features
driving that score vary with the sample. This is expected for
high-dimensional Ridge regression with correlated features and does not
undermine the aggregate CPD results.

\end{document}